\pdfoutput=1

\documentclass[11pt]{article}

\usepackage{acl}

\usepackage{times}
\usepackage{latexsym}
\usepackage{subcaption}
\usepackage[T1]{fontenc}

\usepackage[utf8]{inputenc}

\usepackage{microtype}

\usepackage{inconsolata}
\usepackage{amsmath}
\usepackage{makecell}
\usepackage{multirow}
\usepackage{graphicx}
\usepackage{ulem}
\title{Enhancing Metaphor Detection through Soft Labels and Target Word Prediction}

 \author{Kaidi Jia \and Rongsheng Li\thanks{corresponding author} \\
	College of Computer Science and Technology, Harbin Engineering University, Harbin 150001, China   \\
	dasheng@hrbeu.edu.cn}

\begin{document}
\maketitle
\begin{abstract}
Metaphors play a significant role in our everyday communication, yet detecting them presents a challenge.     Traditional methods often struggle with improper application of language rules and a tendency to overlook data sparsity. To address these issues, we integrate knowledge distillation and prompt learning into metaphor detection.
Our approach revolves around a tailored prompt learning framework specifically designed for metaphor detection. By strategically masking target words and providing relevant prompt data, we guide the model to accurately predict the contextual meanings of these words. This approach not only mitigates confusion stemming from the literal meanings of the words but also ensures effective application of language rules for metaphor detection. Furthermore, we've introduced a teacher model to generate valuable soft labels. These soft labels provide a similar effect to label smoothing and help prevent the model from becoming over confident and effectively addresses the challenge of data sparsity.
Experimental results demonstrate that our model has achieved state-of-the-art performance, as evidenced by its remarkable results across various datasets.

\end{abstract}

\section{Introduction}
Metaphors play a crucial role in our daily lives. By skillfully employing metaphors, expressions become more vivid, concise, and approachable \cite{lakoff2008metaphors}. The accurate identification of metaphors not only advances the field of NLP but also benefits various downstream tasks, including machine translation \cite{shi-2014} and sentiment analysis \cite{dankers-etal-2019-modelling}. Therefore, metaphor detection stands as a pivotal research topic in natural language processing.

The task of metaphor detection poses significant challenges. Early methods \cite{turney-etal-2011-literal, Broadwell2013, tsvetkov-etal-2014-metaphor, bulat-etal-2017-modelling} relied on hand-designed linguistic features to identify metaphors, while later approaches \cite{wu-etal-2018-neural, gao-etal-2018-neural, mao-etal-2019-end} utilized recurrent neural networks (RNNs) to analyze contextual information. However, both categories heavily depended on  crafted structures, and the encoders employed struggled to handle complex contextual information, leading to limited effectiveness. In contrast, more recent methods \cite{gong-etal-2020-illinimet, su-etal-2020-deepmet, choi-etal-2021-melbert, zhang-liu-2022-metaphor, zhou-etal-2023-clcl} leverage pre-trained language models such as BERT \cite{devlin-etal-2019-bert} or RoBERTa \cite{Liu2019RoBERTa} to process contextual information, resulting in significantly improved performance.

While utilizing pre-trained language models has yielded the best results in metaphor detection, these approaches often fail to fully leverage  linguistic rules. Among the most advanced methods \cite{choi-etal-2021-melbert, zhang-liu-2022-metaphor, zhou-etal-2023-clcl}, the incorporation of linguistic rules, such as the Metaphor Identification Procedure (MIP) \cite{Pragglejaz2007}, stands out. MIP operates on the principle of extracting the contextual meaning of a target word in the original sentence. If this contextual meaning diverges from the literal meaning, a metaphorical use is identified. For instance, in the sentence "\textit{We must bridge the gap between employees and management}," the contextual meaning of the target word 'bridge' pertains to reducing differences or divisions, while its literal meaning refers to constructing a physical bridge. The inconsistency between these meanings indicates a metaphorical use. However, despite the logical validity of MIP, current methods fail to effectively utilize it. Typically, these methods \cite{choi-etal-2021-melbert, zhang-liu-2022-metaphor, zhou-etal-2023-clcl} encode the sentence directly using an encoder and extract the word vector from the position of the target word to represent its contextual meaning. Yet, this approach is flawed as the word vector may be influenced by the literal meaning of the target word, leading to an incomplete representation of its contextual meaning.

Furthermore, metaphor detection tasks suffer from severe data sparsity issues. In current datasets, non-metaphorical examples vastly outnumber metaphorical ones. Consequently, using one-hot labels can lead the model to become over-confident \cite{szegedy-2016}, thereby impairing its ability to accurately detect metaphorical usage. Previous methods primarily focused on designing complex structures to maximize the utility of limited data but often overlooked the challenge of data sparsity arising from the imbalance of categories within the dataset.

To address these challenges, we present MD-PK (Metaphor Detection via Prompt Learning and Knowledge Distillation), a novel approach to metaphor detection. By integrating prompt learning and knowledge distillation techniques, our model offers solutions to the issues of improper language rule utilization and data sparsity. Effective utilization of language rules necessitates accurately determining the contextual meaning of target words. To achieve this, we devise a prompt learning template tailored for metaphor detection tasks. By masking the target word in the sentence and providing appropriate hints, our model can generate more contextually relevant words in place of the target word. These generated words serve as the contextual meaning of the target word, thereby significantly reducing interference from its literal meaning. Consequently, MIP language rules can be applied more effectively to detect metaphors. Furthermore, we leverage a teacher model equipped with prior knowledge to generate meaningful soft labels, which guide the optimization process of the student model. Unlike one-hot hard labels, the soft labels produced by the teacher model exhibit properties akin to label smoothing \cite{szegedy-2016}. This helps alleviate the model's tendency towards over-confidence and mitigates the adverse effects of data sparsity. Our work contributions are as follows:
\begin{itemize}
	\item[$\bullet$]We propose a novel metaphor detection module called MIP-Prompt. Through the development of a distinctive prompt learning template tailored specifically for metaphor detection tasks, we facilitate the model in generating contextually relevant words beyond the target word. This approach addresses the challenge of improper utilization of language rules.
	\item[$\bullet$]We incorporate knowledge distillation into the metaphor detection task, enabling the student model to learn from the soft labels generated by the teacher model. This effectively mitigates the model's tendency towards over-confidence and significantly alleviates the data sparsity problem. Additionally, knowledge distillation facilitates rapid acquisition of useful knowledge from the teacher model, thereby enhancing the convergence speed of the student model.
	\item[$\bullet$]Experiments show that our method achieves the best results on multiple datasets. At the same time, we provide detailed ablation experiments and case study to prove the effectiveness of each module.
\end{itemize}

\section{Related Work}
\subsection{Metaphor Detection}
Metaphor usage is pervasive in everyday communication and holds significant importance \cite{lakoff2008metaphors}. Early detection methods relied on extracting linguistic features from corpora  \cite{turney-etal-2011-literal, Broadwell2013, tsvetkov-etal-2014-metaphor, bulat-etal-2017-modelling}. However, these approaches were limited by their dependence on the corpus data. Subsequent studies attempted to leverage recurrent neural networks (RNNs) to extract contextual information for metaphor recognition  \cite{wu-etal-2018-neural, gao-etal-2018-neural, mao-etal-2019-end}, yet struggled to identify complex metaphor usages effectively.

Recent advancements have shifted towards Transformer-based approaches  \cite{gong-etal-2020-illinimet, su-etal-2020-deepmet, choi-etal-2021-melbert, zhang-liu-2022-metaphor, zhou-etal-2023-clcl}. By utilizing pre-trained models, these methods better leverage contextual information, leading to state-of-the-art results. However, despite their success, there remains significant room for improvement due to the neglect of addressing data sparsity issues and the improper utilization of language rules.
\begin{figure}
	\centering
	\includegraphics[width=\linewidth]{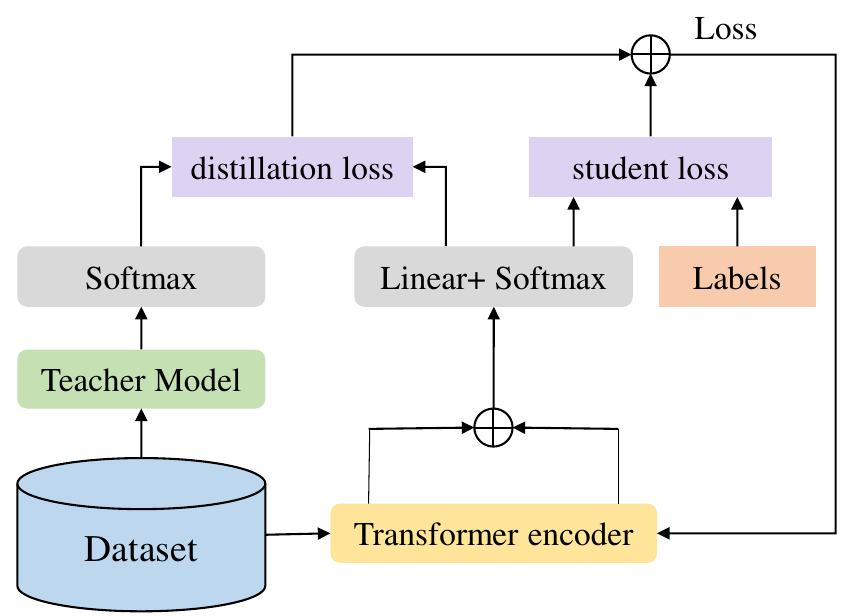}
	\caption{Structrues of MD-PK.}
	\label{fig1}
\end{figure}
\subsection{Knowledge Distillation}
Knowledge distillation, first introduced by \citet{Hinton2015}, aims to improve the performance and accuracy of a smaller model by utilizing supervision from a larger model with superior performance. In this technique, the predictions of the teacher model are referred to as soft labels. \citet{Hinton2015} argued that soft labels, characterized by higher entropy compared to one-hot labels, provide richer information. Mathematical verification by \citet{cheng-2020} demonstrated that using soft labels accelerates the learning process of the student model. Furthermore, \citet{zhao-2022} provided further evidence that the inclusion of non-target category information in soft labels enhances the model's capabilities. Additionally, they established that in scenarios with significant data noise and challenging tasks, the efficacy of knowledge distillation is further enhanced.

\subsection{Prompt Learning}
Prompt learning involves augmenting the input of a downstream task with 'prompt information' to effectively transform it into a text generation task, without substantially altering the structure and parameters of the pretrained language model. Currently, prompt learning has found widespread application in diverse domains such as text classification  \cite{yin-etal-2019-benchmarking}, information extraction  \cite{cui-etal-2021-template}, question answering systems  \cite{khashabi-etal-2020-unifiedqa}, and many others.

In this study, we propose a novel prompt learning template tailored specifically for metaphor detection. By integrating this template into the model architecture, we aim to facilitate the generation of contextual meaning, enabling the model to effectively leverage linguistic rules for enhanced metaphor detection capabilities.

\section{MD-PK}
In this section, we present our proposed model. We begin by outlining the overall structure, which comprises two main components: (1) the metaphor detection module, and (2) the Knowledge distillation module. Subsequently, we delve into each module's functionalities and discuss their roles in the model.
\begin{figure*}
	\centering
	\includegraphics[width=\linewidth]{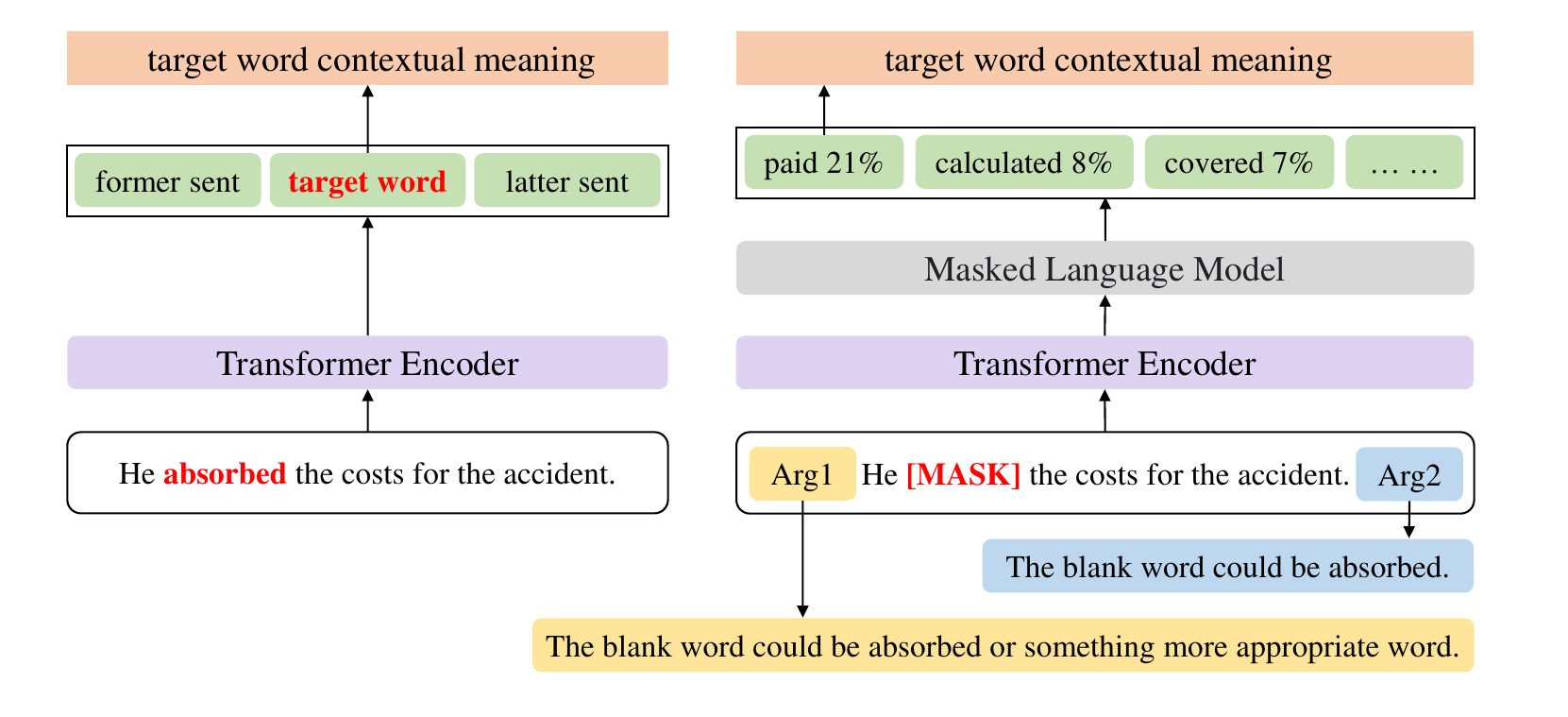}
	\caption{Structures of MIP(left) and MIP-Prompt(right)}
	\label{fig2}
\end{figure*}
\subsection{Overall structure}
Our model, MD-PK, as depicted in Fig. \ref{fig1}, comprises two main components. The first component is the metaphor detection module, responsible for extracting context information from input sentences using an encoder and leveraging linguistic rules to aid metaphor detection. To effectively utilize these language rules, we introduce a prompt learning template. By masking the target word and providing specific prompt information, the model can accurately generate the contextual meaning of the target word, thereby enhancing metaphor detection performance.

The second component is the knowledge distillation module, aimed at facilitating rapid knowledge acquisition by the student model. This is achieved by training the student model to learn from the soft labels generated by the teacher model, which possesses prior knowledge. Additionally, the soft labels act akin to label smoothing, helping mitigate the over-confident tendencies of the student model.

In the knowledge distillation module, the student model corresponds to the metaphor detection model designed in the first component, while the teacher model is a pre-trained model equipped with extensive prior knowledge.

\subsection{Metaphor Detection}
In this paper, we leverage two linguistic rules to aid in metaphor detection: Selectional Preference Violation (SPV) \cite{Yorick1978} and Metaphor Identification Procedure (MIP) \cite{Pragglejaz2007}. SPV operates on the principle of identifying metaphors by discerning inconsistencies between the target word and its context. Conversely, MIP detects metaphors by comparing the contextual meaning of the target word with its literal meaning.

When employing SPV linguistic rules for metaphor detection, we adopt a methodology similar to previous works \cite{choi-etal-2021-melbert, zhang-liu-2022-metaphor, zhou-etal-2023-clcl}. Specifically, We input the sentence containing the target word into an encoder, extract the sentence vector and the target word vector, and input the two vectors as inputs into the full connection layer, through the full connection layer to extract the relationship between the sentence and the target word. Note that the target word here is not replaced with the word generated by the prompt learning.

To apply MIP language rules for metaphor identification, it's crucial to extract the contextual meaning of the target word and juxtapose it with its literal meaning. As illustrated in Fig. \ref{fig2}, previous methodologies have entailed encoding the sentence directly and extracting the vector corresponding to the target word's position in the sentence encoding as its contextual meaning. However, this approach is flawed, as the extracted contextual meaning vector of the target word may be influenced by its literal meaning, thus failing to accurately represent its contextual meaning. 

To extract the true contextual meaning vector of the target word, we introduce a novel MIP module termed MIP-Prompt. In this module, the target word is masked, and prompt information is provided to assist the model in accurately predicting the context meaning of the target word. Specifically, we design a template for predicting the masked word, and utilize the predicted word as the contextual meaning of the target word. By incorporating prompt learning, we ensure that the contextual meaning of the target word remains undisturbed by its literal meaning during extraction, thus enabling more effective utilization of the MIP linguistic rules.

As for the extraction of the literal meaning of the target word, we adopt the method proposed in the previous work \cite{zhang-liu-2022-metaphor} to find out the sentence corresponding to its literal meaning usage for each target word, and put the sentence into the encoder to extract the vector of the position of the target word as the literal meaning of the target word. 

\begin{figure*}
	\centering
	\includegraphics[width=\linewidth]{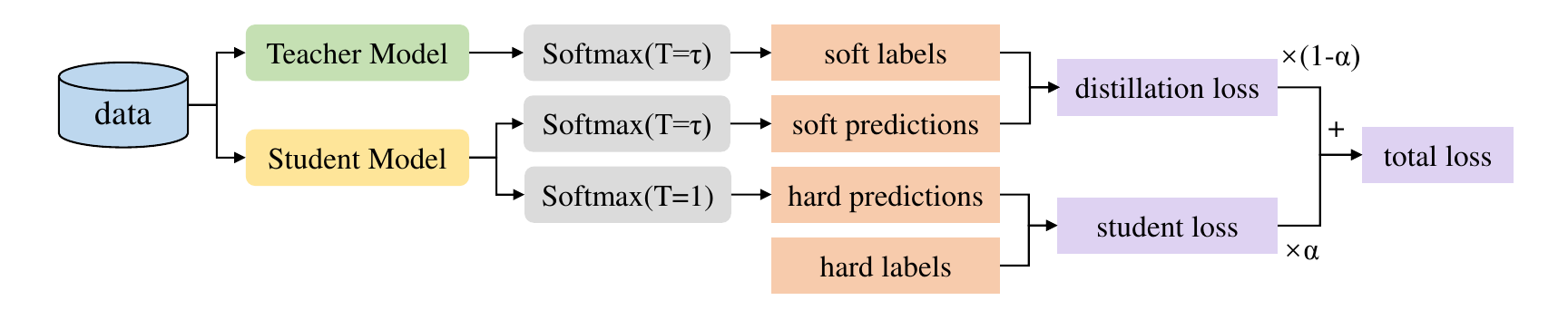}
	\caption{Structures of Knowledge Distillation}
	\label{fig3}
\end{figure*}

\textbf{Template Construction:} Given the original sentence vector $x_s$, we transfer it to $x_{prompt-s}$ using a template:
\begin{equation} 
	\begin{aligned}
		&{{x}_{prompt-s}}=T({{x}_{s1}},{{x}_{s2}})\\&=Arg1+{{x}_{s1}}+[MASK]+{{x}_{s2}}+Arg2
	\end{aligned}
\end{equation}

Where $x_{s1}$ and $x_{s2}$ are sentence vectors before and after the target word, Arg1 and Arg2 are manually designed prompts, and T represents the template function. In this task, we design Arg1 as "The blank word could be [TAR] or something more appropriate word." and Arg2 as "The blank word could be [TAR]." Where [TAR] is the target word vector in the original sentence.

\textbf{Vector Prediction:} We input the designed $x_{prompt-s}$ into RoBERTa \cite{Liu2019RoBERTa} to predict the vector represented by [MASK]. Subsequently, we select the word vector with the highest probability of prediction as the context meaning vector of the target word. This vector is then juxtaposed with the literal meaning vector of the target word to detect metaphors. In all the examples depicted in Fig. \ref{fig2}, the word with the highest probability is "paid," thus we designate "paid" as the contextual meaning of the target word, enabling accurate identification of metaphorical usage.

\subsection{Knowledge Distillation}
Our proposed knowledge distillation method comprises a teacher model T and a student model S. The teacher model T is pre-trained with extensive prior knowledge, enabling it to generate meaningful soft labels to facilitate the optimization process of the student model, and here we use MisNet \cite{zhang-liu-2022-metaphor} as the teacher model. By learning from these soft labels generated by the teacher model, the student model S rapidly enhances its capabilities. Furthermore, given the prevalent issue of data sparsity in the domain of metaphor detection, the introduction of knowledge distillation serves to prevent over-confidence in the model, thereby further enhancing its ability to detect metaphors.

To prevent the student model S from relying excessively on the teacher model and potentially learning incorrect knowledge, it is essential for S to not only align with the soft labels generated by the teacher model but also match the ground-truth one-hot labels during training. As illustrated in Fig. \ref{fig3}, the loss of the model is divided into two components: one calculated using the soft labels generated by both the student and teacher models, and the other calculated using the hard labels generated by the student model and the ground-truth one-hot labels. The total loss of the model is defined as follows.
\begin{equation}
	{{\mathcal{L}}_{s}}=\alpha \mathcal{L}_{hard}+(1-\alpha ){{\mathcal{L}}_{soft}}
\end{equation}
Where $\alpha$ is the balance coefficient. Where $\mathcal{L}_{hard}$ is the ground-truth loss obtained using one-hot hard labels, and ${{\mathcal{L}}_{soft}}$ is the knowledge distillation loss obtained using soft labels. They are defined as follows:
\begin{equation}
	\mathcal{L}_{hard}=-\frac{1}{|N|}\sum\limits_{i=1}^{N}{{{y}_{i}}\log({\hat{y_i}})}
\end{equation}

\begin{equation}
	{{\mathcal{L}}_{soft}}=-\frac{1}{|N|}\sum\limits_{i=1}^{N}{{{p}_{i}}\log({{q_i}})}
\end{equation}

The one-hot vector $y_i$ represents the ground truth label for each instance, whereas the predicted probability distribution is denoted by $\hat{y_i}$. The total count of samples is given by the variable N. The teacher model's pre-softmax logits are expressed as ${Z}_t$, and the student model's equivalent as ${Z}_s$. These logits are transformed into probabilities using the softmax function, normalized by the temperature parameter $\tau$. Specifically, the teacher's model probability is calculated as $\text{${p}$} = \text{softmax} \left( \frac{{Z}_t}{\tau} \right)$, and similarly, the student's model probability is given by $\text{${q}$} = \text{softmax} \left( \frac{{Z}_s}{\tau} \right)$. $\tau$ is the temperature coefficient, which can alleviate the class imbalance problem and help to narrow the gap between the teacher model and the student model.

\section{Experiment}
\subsection{Datasets}
\begin{table}
	\centering
	\resizebox{\linewidth}{!}{
		\begin{tabular}{c|cccc}
			\Xhline{1.2pt}
			\rule{0pt}{12pt}
			\textbf{Dataset} & \textbf{\#Tar} & \textbf{\#M} & \textbf{\#Sent} & \textbf{\#Len}\\
			\hline
			\rule{0pt}{12pt}
			{VUA ALL$_{tr}$} & 116,622 & 11.19 & 6,323 & 18.4\\
			{VUA ALL$_{dev}$} & 38,628 & 11.62 & 1,550 & 24.9\\
			{VUA ALL$_{te}$} & 50,175 & 12.44 & 2,694 & 18.6\\
			\hline
			\rule{0pt}{12pt}
			VUA Verb$_{tr}$ & 15,516 & 27.90 & 7,479 & 20.2\\
			VUA Verb$_{dev}$ & 1,724 & 26.91 & 1,541 & 25.0\\
			VUA Verb$_{te}$ & 5,783 & 29.98 & 2,694 & 18.6\\
			\hline
			\rule{0pt}{12pt}
			MOH-X & 647 & 48.69 & 647 & 8.0\\
			\hline
			\rule{0pt}{12pt}
			TroFi & 3,737 & 43.54 & 3,737 & 28.3\\
			\Xhline{1.2pt}
	\end{tabular}}
	\caption{\label{dataset}
		Datasets information. \textbf{\#Tar}: Number of target words. \textbf{\#M}: Percentage of metaphors. \textbf{\#Sent}: Number of sentences. \textbf{\#Len}: Average sentence
		length.}
\end{table}
As with most works on metaphor detection, we conduct experiments on four widely used public datasets. The statistics of the dataset are shown in Table \ref{dataset}.

\noindent\textbf{VUA ALL} \cite{Steen2010AMF}: The largest metaphor dataset from VUA, which collects data from the BNCBaby corpus. The dataset covers four domains: academic, dialogue, fiction, and news, and contains multiple parts of speech metaphor usages such as verbs, adjectives, and nouns, which has been widely used in metaphor detection work \cite{choi-etal-2021-melbert,zhang-liu-2022-metaphor,zhou-etal-2023-clcl}.

\noindent\textbf{VUA Verb} \cite{Steen2010AMF}: VUA Verb is the verb subset of VUA ALL, containing only metaphorical uses of verbs.

\noindent\textbf{MOH-X} \cite{mohammad-etal-2016-metaphor}: MOH-X collects metaphorical and literal uses of verbs from WordNet, and each word of MOH-X has multiple uses and contains at least one metaphorical use.

\noindent\textbf{TroFi} \cite{birke-sarkar-2006-clustering}: The TroFi dataset contains only metaphorical and literal uses of verbs. The dataset was collected from the Wall Street Journal from 1987-89. We only use the TroFi dataset for zero-shot evaluation.

\subsection{Baselines}
We compare our model with several strong baselines, including RNN-based and Transformer-based models.

\noindent\textbf{RNN\_ELMo} \cite{gao-etal-2018-neural} and \textbf{RNN\_BERT} \cite{devlin-etal-2019-bert}: These two RNN-based sequence labeling models integrate ELMo (or BERT) and GloVe embeddings to encode word representations and employ BiLSTM as their foundational framework.

\noindent\textbf{RNN\_HG} and \textbf{RNN\_MHCA} \cite{mao-etal-2019-end}: RNN\_HG employs MIP to contrast the discrepancies between the literal and contextual senses of the target words, represented by GloVe and ELMo embeddings, respectively. On the other hand, RNN\_MHCA leverages SPV to gauge the distinctions between them and utilizes a multi-head attention mechanism.

\noindent\textbf{MUL\_GCN} \cite{Le_Thai_Nguyen_2020}:MUL\_GCN adopts a multi-task learning framework for both metaphor detection and semantic disambiguation.

\noindent\textbf{MelBERT} \cite{choi-etal-2021-melbert}: The RoBERTa-based model integrates both SPV and MIP architectures for metaphor detection.

\noindent\textbf{MrBERT} \cite{song-etal-2021-verb}: Approach the metaphor detection task by framing it as a relation classification task, utilizing relation embeddings as the input to BERT.

\noindent\textbf{MisNet} \cite{zhang-liu-2022-metaphor}: The RoBERTa-based model leverages both SPV and MIP structures for metaphor detection. Distinguishing itself from MelBERT, it enhances the representation method of the literal meaning of the target word.

	\noindent\textbf{CLCL} \cite{zhou-etal-2023-clcl}:The RoBERTa-based model introduces curriculum learning and contrastive learning for metaphor detection, building upon the foundation laid by MelBERT. In this approach, curriculum learning incorporates the technique of manual evaluation of difficulty to guide the learning process.
\begin{table*}
	\centering
	\resizebox{\linewidth}{!}{
		\begin{tabular}{c|cccc|cccc|cccc}
			\Xhline{1.2pt}
			\rule{0pt}{15pt}
			\multirow{2}{*}{\textbf{Model}}
			& \multicolumn{4}{c|}{\textbf{VUA~ALL}} & \multicolumn{4}{c|}{\textbf{VUA~Verb}} & \multicolumn{4}{c}{\textbf{MOH-X}}\\
			& Acc & P & R & F1 & Acc & P & R & F1 & Acc & P & R & F1\\
			\hline
			\rule{0pt}{15pt}
			\textbf{RNN\_ELMo} \citeyearpar{gao-etal-2018-neural} & 93.1 & 71.6 & 73.6 & 72.6 & 81.4 & 68.2 & 71.3 & 69.7 & 77.2 & 79.1 & 73.5 & 75.6\\
			\textbf{RNN\_BERT} \citeyearpar{devlin-etal-2019-bert} & 92.9 & 71.5 & 71.9 & 71.7 & 80.7 & 66.7 & 71.5 & 69.0 & 78.1 & 75.1 & 81.8 & 78.2\\
			\textbf{RNN\_HG} \citeyearpar{mao-etal-2019-end} & 93.6 & 71.8 & 76.3 & 74.0 & 82.1 & 69.3 & 72.3 & 70.8 & 79.7 & 79.7 & 79.8 & 79.8\\
			\textbf{RNN\_MHCA} \citeyearpar{mao-etal-2019-end} & 93.8 & 73.0 & 75.7 & 74.3 & 81.8 & 66.3 & 75.2 & 70.5 & 79.8 & 77.5 & 83.1 & 80.0\\
			\textbf{MUL\_GCN} \citeyearpar{Le_Thai_Nguyen_2020} & 93.8 & 74.8 & 75.5 & 75.1 & 83.2 & 72.5 & 70.9 & 71.7 & 79.9 & 79.7 & 80.5 & 79.6\\
			\hline
			\rule{0pt}{15pt}
			\textbf{MelBERT}$\dagger$ \citeyearpar{choi-etal-2021-melbert} & 94.0 & 80.5 & \uline{76.4} & \uline{78.4} & 80.7 & 64.6 & \textbf{78.8} & 71.0 & 81.6 & 79.7 & 82.7 & 81.1\\
			\textbf{MrBERT} \citeyearpar{song-etal-2021-verb} & \uline{94.7} & \textbf{82.7} & 72.5 & 77.2 & \uline{86.4} & \textbf{80.8} & 71.5 & \uline{75.9} & 81.9 & 80.0 & \textbf{85.1} & 82.1\\
			\textbf{MisNet}$\dagger$ \citeyearpar{zhang-liu-2022-metaphor} & \uline{94.7} & \uline{82.4} & 73.2 & 77.5 & 84.4 & {77.0} & 68.3 & 72.4 & 83.1 & \uline{83.2} & 82.5 & 82.5\\
			\textbf{CLCL} \citeyearpar{zhou-etal-2023-clcl} & 94.5 & 80.8 & 76.1 & \uline{78.4} & 84.7 & 74.9 & 73.9 & 74.4 & \uline{84.3} & \uline{84.0} & 82.7 & \uline{83.4}\\
			\hline
			\rule{0pt}{15pt}
			\textbf{MD-PK} & \textbf{94.9} & 80.8 & \textbf{77.8} & \textbf{79.3} & \textbf{86.5} & \uline{78.7} & \uline{74.8} & \textbf{76.7} & \textbf{85.6} & \textbf{85.6} & \uline{85.0} & \textbf{85.2}\\
			\Xhline{1.2pt} 
	\end{tabular}}
	\caption{\label{results}
		Results on VUA All, VUA Verb, and MOH-X. Best in bold and second best in italic underlined. The $\dagger$ results are reproduced by \citet{zhou-etal-2023-clcl}}.
\end{table*}

\subsection{Experimental Settings}
In our experiments, we utilize the RoBERTa model \cite{Liu2019RoBERTa} provided by HuggingFace as the encoder. For knowledge distillation, we employ the MisNet model \cite{zhang-liu-2022-metaphor} as the teacher model.

\textbf{VUA ALL}: For VUA ALL dataset, the learning rate is 1e-5 with learning rate warmup, the number of epochs is 10, and the batch size is 64.
\textbf{VUA Verb}: For VUA Verb, the learning rate is 1e-5 and the learning rate warmup is used, the number of epochs is 6, and the batch size is 64.
\textbf{MOH-X}: For MOH-X, the learning rate is fixed to 1e-5, the number of epochs is 15, and the batch size is 64.
\textbf{TroFi}: For TroFi, we only use it for zero-shot evaluation. Where the model is trained on VUA ALL and tested on TroFi.

\subsection{Evaluation Metrics}
In line with previous metaphor detection tasks, we evaluate the model's performance using precision (Acc), precision (P), recall (R), and F1 score (F1). Notably, the F1 score reflects the model's performance specifically regarding the metaphor category, treating it as the positive category.

\section{Results and Analysis}
\subsection{Overall Results}
Table \ref{results} presents the comparative analysis of MD-PK against other robust baseline models across VUA ALL, VUA Verb, and MOH-X datasets. The results underscore the remarkable performance achieved by MD-PK. Specifically, on the VUA-ALL dataset, our model exhibits a notable improvement in F1 score, surpassing RNN-based and Transformer-based models by 7.6 and 2.1, respectively. Compared to the state-of-the-art (SOTA) model CLCL, MD-PK outperforms it by 0.9 F1 score while attaining the highest precision and recall scores. These findings suggest that our model excels in predicting intricate metaphor usage effectively. Furthermore, considering the substantial data sparsity challenge inherent in the VUA ALL dataset, our approach capitalizes on knowledge distillation, leading to significant advantages. This development undoubtedly presents a superior solution to address the data sparsity issue.

On the VUA Verb dataset, our model demonstrates substantial improvements in F1 score compared to RNN-based and Transformer-based models, with enhancements of 7.7 and 5.7, respectively. Moreover, when compared to the state-of-the-art (SOTA) model MrBERT, our model achieves a superior F1 score advantage of 0.8, highlighting its proficiency in predicting the metaphorical usage of verbs. It's noteworthy that MrBERT incorporates a distinctive structure tailored for the verb dataset, leveraging various relations between the subject and object of the verb. In contrast, MD-PK attains superior results solely through semantic matching methods, showcasing the efficacy of the MIP-Prompt module.

On the MOH-X dataset, our model demonstrates significant improvements in F1 score compared to both RNN-based and Transformer-based models, with enhancements of 9.6 and 4.1, respectively. Furthermore, when compared to the state-of-the-art (SOTA) model CLCL, MD-PK exhibits a remarkable advantage with an increased F1 score of 1.8, while also achieving the highest precision and recall scores. These findings underscore the effectiveness of MD-PK in predicting common metaphor usage accurately. It's notable that MOH-X presents the smallest amount of data among the datasets considered. Despite this limitation, our model significantly outperforms all other models on this dataset. This achievement highlights the crucial role of the knowledge distillation module. By leveraging our method, the constraints posed by data scarcity in the field of metaphor detection can be mitigated to a considerable extent.

\begin{table}
	\centering
	\begin{tabular}{c|cccc}
		\Xhline{1.2pt}
		\rule{0pt}{15pt}
		\multirow{2}{*}{\textbf{Model}} &\multicolumn{4}{c}{\textbf{TroFi(Zero-shot)}}\\
		& Acc & P & R & F1\\
		\hline
		\rule{0pt}{15pt}
		\textbf{MelBERT} & - & 53.4 & 74.1 & 62.0\\
		\textbf{MrBERT} & \uline{61.1} & \uline{53.8} & \uline{75.0} & \uline{62.7}\\
		\hline
		\rule{0pt}{15pt}
		\textbf{MD-PK} & \textbf{61.3} & \textbf{54.0} & \textbf{75.4} & \textbf{62.9}\\
		\Xhline{1.2pt}
	\end{tabular}
	\caption{\label{TroFi}
		Zero-shot transfer results on TroFi dataset.}
\end{table}
\subsection{Zero-shot transfer on TroFi}
To assess the model's generalization capability and confirm its efficacy in metaphor detection, we conduct zero-shot transfer experiments on the TroFi dataset. As depicted in Table \ref{TroFi}, our model attains superior results across all metrics, underscoring its robustness across diverse datasets. This outcome serves as compelling evidence that our model exhibits strong generalization ability and is not confined to a specific dataset.

\subsection{Ablation Study}

\begin{table}
	\centering
	\begin{tabular}{c|cccc}
		\Xhline{1.2pt}
		\rule{0pt}{15pt}
		\textbf{Ablation} & \textbf{Acc} & \textbf{P} & \textbf{R} & \textbf{F1}\\
		\hline
		\rule{0pt}{15pt}
		\textbf{-prompt} & 86.0 & 77.3 & \textbf{75.7} & \uline{76.1}\\
		\textbf{-KD} & \uline{86.1} & \uline{78.9} & 73.1 & {75.9}\\
		\textbf{-(prompt\&\&KD)} & {86.0} & \textbf{79.8} & 71.2 & 75.2\\
		\hline
		\rule{0pt}{15pt}
		\textbf{MD-PK} & \textbf{86.4} & 78.7 & \uline{74.8} & \textbf{76.7}\\
		\Xhline{1.2pt}
	\end{tabular}
	\caption{\label{Effective}
		Effectiveness study on VUA Verb dataset.}
\end{table}

To examine the impact of different components of our approach, namely MIP-Prompt and knowledge distillation, we evaluated variants without prompt learning (-prompt), without knowledge distillation (-KD), and the original model trained with identical hyperparameters on the VUA Verb dataset. As illustrated in Table \ref{Effective}, the absence of prompt learning or knowledge distillation leads to a decrease in both accuracy and F1 score to some extent.

The improvement effect observed with prompt learning and the knowledge distillation module closely mirrors that of the original model, underscoring the indispensable role of both modules. However, it is only through the combined utilization of prompt learning and knowledge distillation that they can synergistically complement each other, resulting in optimal performance.

\begin{figure}
	\centering
	\includegraphics[width=\linewidth]{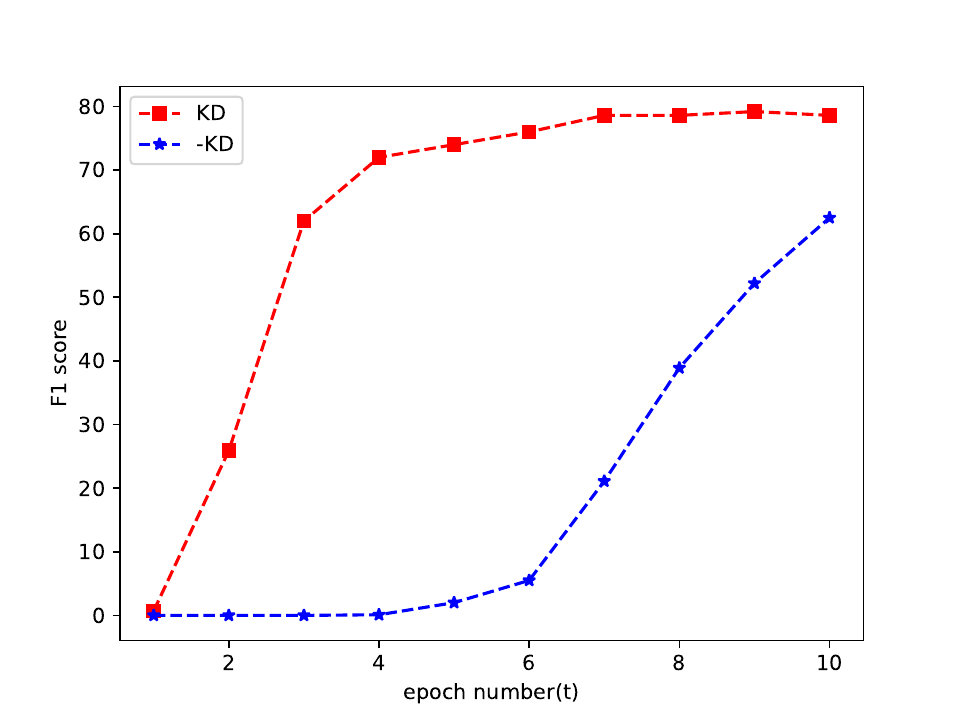}
	\caption{Visualization of convergence rates of different structures}
	\label{fig5}
\end{figure}

\subsection{Analysis on Knowledge Distillation}
Following the introduction of knowledge distillation, the student model adeptly assimilates the knowledge from the teacher model, resulting in rapid enhancement of its capabilities. As illustrated in Fig. \ref{fig5}, in the absence of knowledge distillation, the model struggles to acquire meaningful features during the initial stages of training. Moreover, exacerbated by the severe data sparsity issue, the model tends to exhibit over-confidence, leading to disproportionately low F1 scores despite high accuracy rates.

Conversely, with knowledge distillation in place, the student model swiftly acquires knowledge from the teacher model. Furthermore, leveraging the richer information encapsulated in the soft labels generated by the teacher model helps alleviate the issue of over-confidence in the student model, thereby significantly accelerating convergence speed.

In addition, we have studied the effects of parameters in knowledge distillation, more details can be found in the appendix \ref{kd}.

\subsection{Analysis on Prompt Learning}
While the incorporation of prompt learning enhances the precision with which the contextual significance of a target word is represented, there is a potential risk of substituting the target word with one that bears no relevance to the source material, thereby undermining the model's predictive capabilities. As illustrated in Table \ref{Effective}, the elimination of prompt learning results in a decline in the model's precision, accompanied by a corresponding increase in recall. This suggests that prompt learning primarily refines the model's accuracy in forecasting positive classes, without necessarily facilitating the model's ability to misclassify negative instances as positive. Consequently, prompt learning is not inclined to treat the target word as a context-independent entity.

\section{Conclusion}
In this paper, we propose MD-PK, a novel metaphor detection model comprising two key modules: metaphor detection and knowledge distillation. By integrating knowledge distillation and prompt learning into the realm of metaphor detection, we effectively address challenges associated with improper utilization of language rules and data sparsity. Our method is evaluated across four datasets, showcasing considerable improvements over strong baseline models. Furthermore, detailed ablation experiments are conducted to elucidate the efficacy of our approach.

\section*{Limitations}
When predicting the contextual meaning, we utilize the RoBERTa model. However, the inherent limitations of the RoBERTa model may hinder its ability to predict the most suitable contextual meaning. By incorporating Large Language Models (LLMs), we anticipate further enhancements in model performance. Additionally, our designed prompt learning template may inadvertently increase the likelihood of predicting the literal meaning of the target word. Therefore, designing a more suitable prompt learning template remains an avenue for future research.

\bibliography{anthology,custom}

\appendix
\clearpage

\begin{figure}
	\centering
	\subfloat[Mean Acc scores]  
	{
		\label{fig:subfig1}\includegraphics[width=\linewidth]{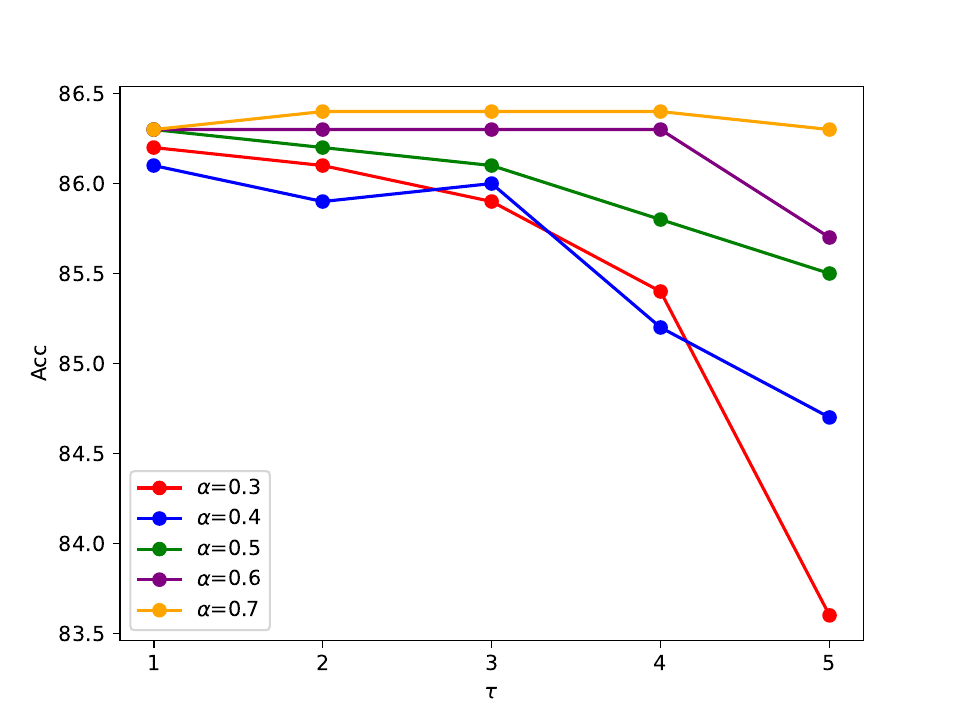}
	}
	\\
	\subfloat[Mean F1 scores]
	{
		\label{fig:subfig2}\includegraphics[width=\linewidth]{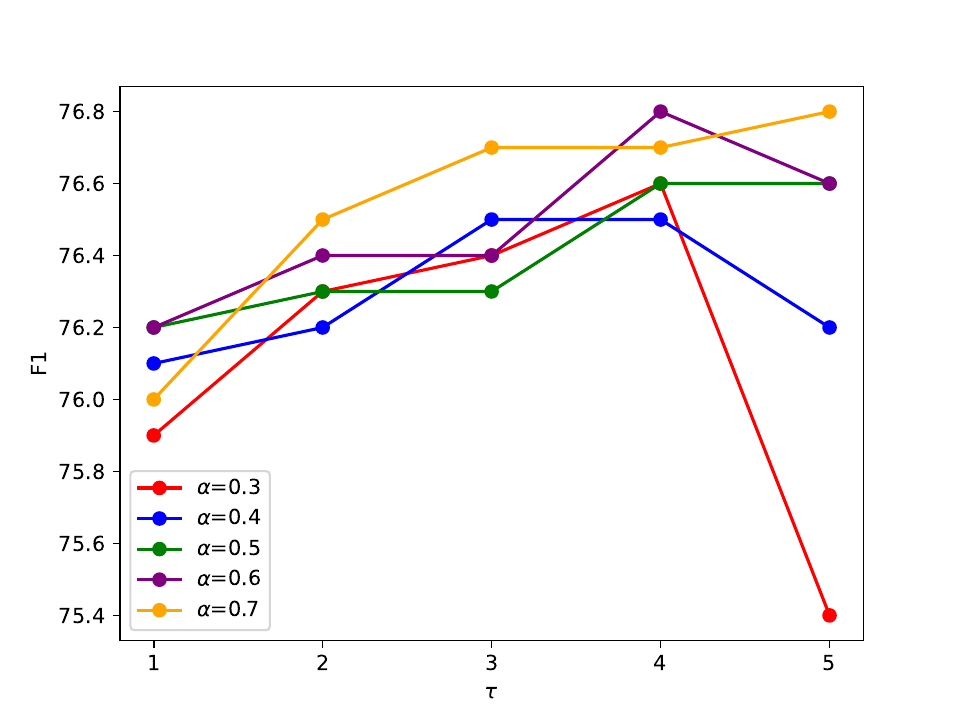}
	}
	\caption{Mean Acc./F1 scores of different hyparameters on VUA Verb dataset.}
	\label{fig4}
\end{figure}

\section{\label{kd}Influence of Hyperparameter in KD}
Given the sensitivity of the knowledge distillation algorithm to hyperparameters, we conducted experiments on the VUA Verb dataset, varying $\alpha$ from 0.3 to 0.7 and $\tau$ from 1 to 5. As depicted in Fig. \ref{fig4}, the average performance notably improves with larger values of $\alpha$, suggesting that the enhancement in model capability primarily relies on one-hot hard labels. Additionally, as $\tau$ increases, the model's performance generally exhibits an initial rise followed by a decline. This trend indicates that $\tau$ should be appropriately increased to enable the student model to effectively assimilate the latent knowledge contained in soft labels. However, excessive values of $\tau$ should be avoided to mitigate the adverse impact of negative labels.

\end{document}